\documentclass[10pt,letterpaper]{article}
\usepackage[top=0.85in,left=2.75in,footskip=0.75in]{geometry}
\usepackage{comment}
\usepackage{amsmath,amssymb}
\usepackage{todonotes}
\usepackage{booktabs}
\usepackage{changepage}
\usepackage{threeparttable}
\usepackage{textcomp,marvosym}

\usepackage{cite}

\usepackage{nameref,hyperref}

\usepackage[right]{lineno}

\usepackage[nopatch=eqnum]{microtype}
\DisableLigatures[f]{encoding = *, family = * }

\usepackage{array}
\usepackage{longtable}
\usepackage{array}

\newcolumntype{+}{!{\vrule width 2pt}}

\newlength\savedwidth

\raggedright
\usepackage[aboveskip=1pt,labelfont=bf,labelsep=period,justification=raggedright,singlelinecheck=off]{caption}

\makeatletter
\renewcommand{\@biblabel}[1]{\quad#1.}
\makeatother

\usepackage{lastpage,fancyhdr,graphicx}
\usepackage{epstopdf}
\fancyheadoffset[L]{2.25in}
\fancyfootoffset[L]{2.25in}
\begin{document}
\vspace*{0.2in}

\begin{flushleft}
{\Large
\textbf\newline{Clinical Insights: A Comprehensive Review of Language Models in Medicine.} 
}
\newline
\\
Nikita Neveditsin\textsuperscript{1*},
Pawan Lingras\textsuperscript{1},
Vijay Mago\textsuperscript{2},
\\
\bigskip
\textbf{1} Department of Mathematics and Computing Science, Saint Mary's University, Halifax, Nova Scotia, Canada
\\
\textbf{2} School of Health Policy and Management, York University, Toronto, Ontario, Canada
\\

\bigskip

%
%



* nikita.neveditsin@smu.ca

\end{flushleft}
\section*{Abstract}
\color{black}

This paper explores the advancements and applications of language models in healthcare, focusing on their clinical use cases. It examines the evolution from early encoder-based systems requiring extensive fine-tuning to state-of-the-art large language and multimodal models capable of integrating text and visual data through in-context learning. The analysis emphasizes locally deployable models, which enhance data privacy and operational autonomy, and their applications in tasks such as text generation, classification, information extraction, and conversational systems. The paper also highlights a structured organization of tasks and a tiered ethical approach, providing a valuable resource for researchers and practitioners, while discussing key challenges related to ethics, evaluation, and implementation.

\color{black}


\section{Introduction}
The advances in the area of artificial intelligence (AI) in recent years has opened countless opportunities for various sectors, including healthcare. The potential influence of AI is a subject of debate concerning its impact on humanity. Leading AI experts have called for caution, evidenced by an open letter urging a pause in the expansion of advanced AI models, which reflects growing concerns among policymakers and the public about the ethical, social, and economic ramifications of AI. While some argue that AI can bring substantial advances in efficiency and effectiveness across many sectors, others fear it could exacerbate inequalities, displace jobs, and challenge societal norms \cite{Goldfarb}. While AI in healthcare has an extensive history of research \cite{Secinaro2021}, the emergence of advanced foundational models \cite{bommasani2021opportunities} such as GPT family\cite{openai2023gpt4}, Gemini \cite{geminiteam2024}, and a series of open models like the Llama \cite{touvron2023llama2}, offers unprecedented perspectives for the transformation of the healthcare sector.

\color{black}
This review paper synthesizes and critically examines the landscape of language models in the medical domain, with a particular emphasis on their clinical applications. Compared to other reviews in this area, it highlights the following aspects:
\begin{enumerate}
    \item A dedicated focus on clinical applications.
    \item An emphasis on locally deployable models.
    \item A structured organization based on NLP task categories, with appropriate justification for this framework.
    \item A structured ethical assessment of reviewed tasks.
    \item A detailed discussion of evaluation challenges.
\end{enumerate}

Table~\ref{tab:review_comparison} outlines how these distinguishing aspects compare to the focus areas of other reviews in the field.
\begin{table}[ht]
\begin{adjustwidth}{-2.25in}{0in}
\centering
\caption{Unique aspects of this review compared to others in the field.}
\label{tab:review_comparison}
\begin{tabular}{|p{6cm}|p{2cm}|p{2cm}|p{2cm}|p{2cm}|p{2cm}|}
\hline
\textbf{Criterion} & \textbf{Luo et al. (2024) \cite{LUO2024102904}} & \textbf{Meng et al. (2024) \cite{meng2024application}} & \textbf{Wang et al. (2023) \cite{wang2023pre}} & \textbf{Thirunav- ukaras et al. (2023) \cite{Thirunavukarasu2023}} & \textbf{Yang et al. (2023) \cite{yang2023large}} \\ \hline
Focus on clinical applications & Partially & Partially & No & Yes & Yes \\ \hline
Emphasis on locally deployable models & Partially & No & Yes & No & Partially \\ \hline
Structured by NLP task categories with justification & Yes & Partially & Yes & No & Partially \\ \hline
Structured ethical assessment of reviewed tasks & No & Partially & No & Partially & Partially \\ \hline
Detailed discussion of evaluation challenges & Partially & Partially & Partially & Partially & Partially \\ \hline
\end{tabular}
\end{adjustwidth}
\end{table}

This review navigates the advancements from early encoder-based systems to state-of-the-art large language and multimodal models. It places a particular emphasis on locally deployable models, which enhance data privacy and adaptability within healthcare settings. Additionally, the structured exploration of ethical considerations and evaluation challenges provides a critical perspective on the current state of these models. By offering both a clear roadmap and valuable insights, this review serves as a comprehensive resource for newcomers and experienced interdisciplinary researchers, fostering innovation and promoting responsible implementation in this vital domain.

\color{black}

\subsection{Definitions}

\hspace*{\parindent}\textbf{Artificial Intelligence (AI):} A multidisciplinary field aimed at creating systems capable of tasks that typically require human intelligence, such as problem-solving, learning, reasoning, and perception \cite{russell2016artificial}.

\textbf{Machine Learning (ML):} A branch of AI focused on developing algorithms and statistical models that enable computer systems to improve performance on specific tasks by learning from data and identifying patterns, rather than relying on explicit programming \cite{russell2016artificial, ibm_ai_ml_dl_nn}.  

\textbf{Natural Language Processing (NLP):} A branch of AI that deals with the interaction between computers and human language, allowing machines to understand, interpret, and generate human languages \cite{russell2016artificial, jm3}.

\textbf{Language Models:} Probabilistic models that learn the statistical structure of word sequences, enabling the prediction or generation of text based on patterns observed in previously seen data \cite{jm3}.

\textbf{Large Language Models (LLMs):} Language models trained on extensive text corpora using deep learning architectures, enabling them to acquire broad linguistic knowledge and perform diverse natural language tasks \cite{roberts-etal-2020-much, jm3}.

\textbf{Generative Large Language Models (Generative LLMs)}: A subclass of LLMs that generate coherent and contextually relevant text by learning statistical patterns from large-scale language datasets, enabling tasks such as text generation, summarization, and dialogue creation.

\textbf{Open LLMs:} LLMs developed and distributed with open access to their architectures and weights, facilitating transparency, reproducibility, and community-driven improvements \cite{touvron2023llama}.

\textbf{Foundation Models:} Large-scale, general-purpose AI models, such as LLMs, trained on massive and diverse datasets. They serve as a base for a wide range of downstream tasks, often requiring minimal additional training or fine-tuning \cite{bommasani2021opportunities}.

\textbf{Embeddings:} Dense vector representations of words, phrases, or other linguistic units that capture semantic relationships and similarity based on contextual usage in large corpora \cite{jm3}.

\textbf{API (Application Programming Interface):} In the context of LLMs, an API provides a set of tools and protocols that enable users to interact with a language model remotely, without requiring local deployment. Typically, API access abstracts away the model's internals, offering only input-output functionality.

\textbf{GPU (Graphics Processing Unit):} A specialized processor originally designed for rendering graphics, now widely used to accelerate parallelizable computations, particularly in AI and ML tasks. Locally deployed models often require GPUs to run efficiently due to their ability to handle large-scale matrix operations.

\color{black}

\section{Background}
\subsection{Language Models}
\color{red}

\color{black}
A critical milestone in the field of Natural Language Processing (NLP) was the introduction of the attention mechanism in neural machine translation \cite{DBLP:journals/corr/BahdanauCB14}. This mechanism, linking the encoder and decoder in sequence-to-sequence models, paved the way for subsequent advancements by enabling the model to focus on different parts of the input sequence for each step of the output, substantially improving the handling of longer input sequences and complex dependencies. Notably, it led to the creation of the Transformer model \cite{DBLP:journals/corr/VaswaniSPUJGKP17}, which exclusively relies on attention mechanisms. This innovation revolutionized not only the field of NLP but also the broader realms of AI and machine learning (ML).

The emergence of Transformer-based models has spurred the development of a wide array of \textit{language models}, both in commercial and open formats. Among these, the GPT series \cite{Radford2019LanguageMA} has advanced text generation capabilities while establishing the foundation for interactive applications. A notable milestone was the introduction of ChatGPT \cite{OpenAI2023ChatGPT}, a conversational agent that revolutionized chat-based language models by facilitating intuitive, human-like dialogue generation. The success of ChatGPT and similar systems \cite{geminiteam2024, claude3_model_card} underscored the adaptability of Transformer architectures and catalyzed the development of language models optimized for interactive and specialized applications, including those in the medical domain.

The overall task of language modeling can be expressed as estimating the joint probability of a sequence of words $w_1, w_2,\ldots,w_n$ in a sentence, drawn from large text corpora:

\[
P(w_1, w_2, \ldots, w_n) = \prod_{i=1}^{n}P(w_i \mid w_1, w_2, \ldots, w_{i-1})
\]

Despite the simplicity of statistical approaches to language modeling and their initial lack of attention to the underlying rules of language, as critiqued by Chomsky \cite{norvig2017chomsky}, contemporary large language models (LLMs) like GPT-4 have demonstrated remarkable proficiency in a wide range of language understanding tasks \cite{openai2023gpt4}, exhibiting emergent abilities and an advanced capacity to learn natural language patterns.

\color{black}
Modern language models have undergone a notable paradigm shift, moving from a pre-training and fine-tuning approach to embracing in-context learning (ICL) \cite{liu_deid-gpt_2023} and zero-shot learning \cite{DBLP:journals/corr/abs-2005-14165}. Traditionally, language model development has followed a two-step process: pre-training and fine-tuning. During pre-training, a model learns general language representations from large text corpora in an unsupervised manner. For example, BERT \cite{bert} predicts masked tokens, while GPT-2 \cite{Radford2019LanguageMA} predicts the next token in a sequence. The fine-tuning phase then adapts these pre-trained models to specific tasks (e.g., classification, summarization) using labeled data. Although effective, this approach often demands substantial task-specific datasets, which can limit its scalability for diverse downstream applications.

In contrast, zero-shot learning enables models to generalize to unseen tasks \textit{without requiring task-specific fine-tuning}. Instead, the model leverages its pre-trained knowledge and interprets carefully designed prompts to perform tasks directly. For example, a zero-shot model can respond to a prompt such as ``Summarize the key findings of this medical report" without being explicitly trained on clinical summarization datasets.

In-context learning enhances the capabilities of these models by enabling them to solve tasks using information provided within the input prompt. Unlike fine-tuning, ICL does not require updating the model's weights. Instead, the model temporarily ``learns" from examples in the prompt. For instance, the model can be provided with a series of medical cases that include patient symptoms, relevant background knowledge, and corresponding diagnoses. When presented with a new case, such as a patient with specific symptoms (e.g., coughing up phlegm and blood), the model can infer the most likely diagnosis based on the examples and background knowledge, without explicit task-specific fine-tuning \cite{wularge}.

Nevertheless, challenges remain, as highlighted by Mahowald et al. \cite{mahowald2023dissociating}, who emphasize the gap between formal and functional linguistic competencies in LLMs. Formal linguistic competencies refer to a model's ability to understand and generate syntactically and semantically correct language, enabling tasks like sentence completion, grammar correction, or summarization. In contrast, functional competencies involve applying language in practical, goal-oriented contexts, such as interpreting patient symptoms to suggest a diagnosis or deriving actionable steps from clinical guidelines. Achieving functional performance requires not only linguistic fluency but also \textit{reasoning} and task-specific understanding, integrating broader contextual and intentional knowledge.

This gap can be analogized to human neuroscience, where distinct neural mechanisms underpin linguistic processing and reasoning \cite{mahowald2023dissociating}. Similarly, LLMs, trained predominantly on textual data, excel at linguistic mimicry but often lack the modularity and cognitive integration necessary for complex functional reasoning. To bridge this dissociation, Mahowald et al. propose integrating modular architectures or revising training processes to better align with these competencies.

For medical professionals, this gap underscores the limitations of LLMs in healthcare. While LLMs may summarize medical literature with linguistic precision, they may struggle to apply this information effectively in clinical scenarios, such as tailoring treatment plans based on patient-specific data. Recognizing these limitations is critical to ensuring LLMs are deployed as tools that complement, rather than replace, human expertise.

\color{black}

Notably, current research is shifting towards multimodal models that integrate multiple \textit{modalities}, such as visual and textual data, into a single model \cite{openai2023gpt4, geminiteam2024, chameleonteam2024chameleonmixedmodalearlyfusionfoundation}, paving the way for more comprehensive AI solutions. While this review focuses on text-only language models, it briefly examines the ecosystem of multimodal models to highlight their emerging relevance. This exploration provides readers with starting points for further investigation into how these models expand the scope of AI applications, particularly in domains like medical imaging, where integrating textual and visual data is essential.

\subsection{Locally-Deployable Models}

\color{black}

Models can be categorized based on their data management and deployment strategies: locally-deployable models and API-based models. API-based models, such as GPT-4, require data transfer to third-party servers via web interfaces or APIs. In contrast, locally-deployable models run on an organization’s hardware, offering full data control and independence from external vendors.

In the medical domain, locally-deployable models provide critical advantages by ensuring sensitive medical data remains internal, complying with stringent data protection regulations, and enhancing operational autonomy by eliminating reliance on third-party providers.

Many locally-deployable models feature open weights and permissive usage terms \cite{grattafiori2024llama3herdmodels}. This flexibility allows them to be customized for medical domain, optimized for specific types of inquiries, and seamlessly integrated with confidential datasets, while maintaining strict patient confidentiality.

\begin{table}[h!]
\footnotesize
\centering
\begin{threeparttable}
\caption{Distribution of Widely Used Pre-trained Open-Source General-Purpose Models by Number of Parameters and Architecture}
\begin{tabular}{l>{\raggedright\arraybackslash}p{3cm}>{\raggedright\arraybackslash}p{3cm}>{\raggedright\arraybackslash}p{2.6cm}>{\raggedright\arraybackslash}p{2.2cm}}
\toprule
\textbf{Size} & \textbf{Decoder-based} & \textbf{Encoder-based} & \textbf{Encoder-Decoder} & \textbf{Multimodal} \\
\midrule
$\le$ 1B & GPTNeo\cite{gpt-neo} (125-350M) & BERT\cite{bert} (110-340M), ALBERT\cite{lan2020albertlitebertselfsupervised} (12-235M), DeBERTa \cite{deberta} (134M), ELECTRA \cite{clark2020electra} (14-335M), RoBERTa \cite{roberta} (125-355M) & BART\cite{bart}, BertGeneration\cite{Rothe_2020} (140-400M), Flan-T5 \cite{flant5} (77-783M), Pegasus \cite{zhang2019pegasus} (568M), T5 \cite{t5} (60-770M) & BLIP-2 \cite{li2022lavis} (188M), CLIP \cite{radford2021learningtransferablevisualmodels} (428M), deplot \cite{liu2023deplotoneshotvisuallanguage} (300M), Donut \cite{donut} (200M), LayoutLMv3 \cite{huang2022layoutlmv3pretrainingdocumentai} (133-368M) \\

$\le$  10B & CTRL\cite{ctrl} (1.63B), Falcon\cite{falcon} (7B) GPT-J\cite{gpt-j} (6B), Gemma \cite{gemma} (2-7B), Llama \cite{touvron2023llama, touvron2023llama2} (7B, 8B), Mistral \cite{mistral_7b} (7B), Phi-3 \cite{abdin2024phi3technicalreporthighly} (3.8B) & DeBERTa (1.5B) & Flan-T5 (3B), LongT5 \cite{longt5} (3B), T5 (3B) & Fuyu \cite{adept_fuyu8b} (8B), BLIP-2-Opt (3.8B), Llava \cite{llava} (7B), PaliGemma \cite{beyer2024paligemmaversatile3bvlm} (3B),  Chameleon \cite{chameleonteam2024chameleonmixedmodalearlyfusionfoundation} (7B), LLaVa \cite{llava} (7-8B)\\

$\le$  20B & GPT-NeoX \cite{gpt-neox-20b} (20B), Llama (13B) & N/A & Flan-T5 (11B), T5 (11B), UL2 \cite{ul2} (20B) \\

$\le$  80B & Cohere \cite{cohere_command} (35B), Falcon (40B), Llama (65B, 70B) & N/A & N/A & Chameleon (34B), LLaVa (34B)  \\

$\le$  80B & DBRX\cite{databricks_dbrx} (132B), Falcon (180B), OPT \cite{opt} (175B) & N/A & N/A & BLOOM \cite{bloom} (176B)\\

\bottomrule
\end{tabular}
\begin{tablenotes}
\footnotesize
\item Note: B is billions, M is millions.
\end{tablenotes}

\label{tab:llms_by_params}
\end{threeparttable}

\end{table}

The primary limitations of locally-deployable models are their substantial hardware requirements and the technical expertise needed for deployment. Advanced models often comprise a significant number of parameters, typically represented as floating-point values. While a higher parameter count generally improves reasoning, pattern recognition, and linguistic capabilities \cite{JMLR:v25:23-0870}, it also increases storage demands and necessitates more powerful GPUs for both inference and fine-tuning.

Recent advancements have addressed the computational challenges of large models through parameter-efficient fine-tuning and memory optimization. Techniques like Low-Rank Adaptation (LoRA) \cite{hu2021lora} update only low-rank parameters, reducing the need to adjust all model weights. Quantization \cite{jacob2017quantizationtrainingneuralnetworks} further minimizes memory usage and accelerates inference, enabling broader hardware compatibility. Combining these, Quantized Low-Rank Adaptation (QLoRA) \cite{qlora} efficiently fine-tunes large models while significantly lowering memory and computational requirements. For example, a 1-billion-parameter model using 16-bit precision requires around 2GB of GPU memory, whereas quantization can halve this demand. Additional innovations, such as Virtual Memory Stitching (VMS) \cite{10.1145/3620665.3640423}, optimize GPU memory allocation, reducing fragmentation and usage. Flash Attention \cite{flashattention} further enhances hardware efficiency, expanding the feasibility of deploying larger models on on-premise systems.

\color{black}
Table~\ref{tab:llms_by_params} summarizes widely used locally-deployable textual and multimodal models (predominantly English-language) categorized by the number of parameters and their architectures. \color{black}
The table distinguishes between three primary architectures: encoder-only, decoder-only, and encoder-decoder, each of which brings its own strengths and use cases. 

Encoder-only models (e.g., BERT \cite{bert}, DeBERTa \cite{deberta}) excel in understanding and classification tasks, consistently achieving strong performance on comprehension-oriented benchmarks \cite{rajpurkar2016squad, lai2017race, wang2019superglue}. Decoder-only models (e.g., GPT-Neo \cite{gpt-neo}, Falcon \cite{falcon}) are better suited for generative tasks, such as text completion and narrative generation. Encoder-decoder models (e.g., BART \cite{bart}, Flan-T5 \cite{flant5}) strike a balance between comprehension and generation. While multimodal models can extend the capabilities of any of these architectures by incorporating multiple modalities (e.g., text and vision), we deliberately omit these details to maintain the review’s focus on text-based systems.

\subsubsection*{General Performance Guidelines}

General evaluation of encoder-only language models typically focuses on tasks that assess deep language understanding without requiring generative capabilities. Benchmarks such as GLUE \cite{wang2018glue} and SuperGLUE \cite{wang2019superglue}, encompassing tasks like natural language inference, sentiment analysis, paraphrase detection, and coreference resolution, are widely adopted. Many of these benchmarks maintain leaderboards to track state-of-the-art performance \cite{glue_leaderboard}. Since encoder-only models often need fine-tuning for specific downstream tasks, it is advisable to prioritize evaluation on benchmarks that reflect the intended application.

The performance of decoder-only models is typically measured by their ability to generate coherent text, comprehend and reason about given inputs, and follow instructions. Prominent benchmarks include MMLU \cite{hendrycks2020measuring} and MMLU-Pro \cite{wang2024mmlu}, while the technical reports of models like GPT-3 \cite{DBLP:journals/corr/abs-2005-14165} provide a comprehensive list of metrics tailored for generative, decoder-only models.

Encoder-decoder models, designed for sequence-to-sequence tasks, are usually evaluated on benchmarks that measure their capacity to transform input sequences into output sequences. Examples include SQuAD \cite{rajpurkar2016squad} for question answering and XSum \cite{narayan2018dontdetailsjustsummary} for summarization. Readers interested in a deeper understanding of encoder-decoder architectures and their performance considerations are encouraged to review the T5 paper \cite{t5}.

Evaluation of multimodal models often involves visual question answering (VQA) benchmarks. For instance, TextVQA \cite{singh2019vqamodelsread} focuses on text reading in natural images, DocVQA \cite{mathew2021docvqadatasetvqadocument} targets document understanding, and ChartQA \cite{masry2022chartqabenchmarkquestionanswering} assesses chart comprehension. The Gemini technical report \cite{geminiteam2024} offers a comprehensive list of metrics for vision-and-language multimodal models.

\color{black}

\subsection{Domain-specific Language Models in Healthcare}

\begin{table}
\footnotesize
\begin{adjustwidth}{-2.25in}{0in}
    \caption{Overview of Prevalent Medical Language Models}
    \centering
    \begin{tabular}{p{2.5cm}p{1cm}p{1cm}p{5.5cm}p{6cm}}
\toprule
Name & Year (Appx.) & Arch. & Training Data & Experimental Datasets \\
\midrule
ClinicalBERT \cite{huang2020clinicalbert} & 2019 & BERT & MIMIC-III \cite{johnson2016mimiciii}& MIMIC-III \\

BioBERT \cite{Lee_2019} & 2019 & BERT & PubMed abstracts and PMC full-text articles & NER: NCBI \cite{dougan2014ncbi}, i2b2/VA \cite{i2b2}, BC5\cite{li2016biocreative}, BC4CHEMD \cite{krallinger2015chemdner}, BC2GM \cite{smith2008overview}, JNLPBA \cite{huang2019revised}, LINNAEUS \cite{gerner2010linnaeus}, Species-800 \cite{pafilis2013species}; RE: GAD \cite{bravo2015extraction}, EU-ADR \cite{coloma2011combining}, CHEMPROT \cite{kringelum2016chemprot}; QA:  BioASQ \cite{tsatsaronis2015overview} 4b, 5b, 6b, and 7b. \\

BiomedBERT \cite{chakraborty-etal-2020-biomedbert} & 2020 & BERT & Pre-trained on the BREATHE dataset \cite{chakraborty-etal-2020-biomedbert} & NER: NCBI, BC5CDR, BC4CHEMD, BC2GM, JNLPBA; RE: GAD and EU-ADR; QA: SQuAD \cite{rajpurkar2016squad} v1.1 and v2.0, BioASQ 4b, 5b, 6b, and 7b. \\

PubMedBERT \cite{Gu_2021} & 2020 & BERT & PubMed only & BC5, NCBI, BC2GM, JNLPBA, EBM PICO \cite{nye2018corpus}, CHEMPROT, DDI \cite{herrero2013ddi}, GAD, BIOSSES \cite{souganciouglu2017biosses}, HoC \cite{hanahan2000hallmarks}, PubMedQA \cite{PubMedQA}, BioASQ \\

BEHRT \cite{li2020behrt} & 2020 & BERT & Clinical Practice Research Datalink (CPRD) \cite{herrett2015data} & Predict diseases in future patients' visits with CPRD \\

GatorTron \cite{yang_gatortron_2022} & 2022 & BERT-style & UF Health \cite{ufhealth_idr}, PubMed articles, Wikipedia & Clinical Concept Extraction: i2b2, n2c2 \cite{n2c2}; RE: n2c2; STS: n2c2/OHNLP Clinical STS; NLI: MedNLI \cite{romanov2018lessons}; QA: emrQA \cite{pampari2018emrqa}. \\

BioGPT \cite{luo_biogpt_2022} & 2022 & GPT-2 XL & 15 million PubMed items & RE: BC5CDR, KD-DTI \cite{hou2022discovering}, DDI \cite{herrero2013ddi}. MQA: PubMedQA. Document classification: HoC. Text generation: custom dataset \cite{luo_biogpt_2022}.  \\

ClinicalT5 \cite{lu_clinicalt5_2022} & 2022 & T5 & MIMIC-III (textual notes) & Document classification: HoC; NER: NCBI \cite{dougan2014ncbi}, BC5CDR-disease; NLI: MedNLI; Real-world evaluation based on MIMIC-III to predict ICU readmission and mortality risks \\

AlpaCare \cite{zhang_alpacareinstruction-tuned_2023} & 2023 & LLaMA with IFT & MedInstruct-52 (introduced in the paper): 52000 instruction-response pairs generated using GPT-4 prompted with a clinician-crafted seed set & iCliniq2  \cite{zhang_alpacareinstruction-tuned_2023}, MedInstruct-test \\

BioInstruct \cite{tran_bioinstruct_2023} & 2023 & LLaMA with IFT & BioInstruct dataset (introduced in the paper): 25000 instructions in natual language collected with GPT4 & QA: MedQA-USMLE \cite{jin2021disease}, MedMCQA \cite{pal2022medmcqa}, PubMedQA, BioASQ MCQA; NLI: MedNLI. Text Generation: Conv2note \cite{conv2note}, ICliniq \cite{icliniq}. \\

ChatDoctor \cite{li_chatdoctor_2023} & 2023 & LLaMA & Conversations from HealthCareMagic-100k \cite{li_chatdoctor_2023} & HealthCareMagic100k, iCliniq  \\

Clinical Camel \cite{toma_clinical_2023} & 2023 & LLaMA & Clinical articles converted into synthetic dialogues, data from ShareGPT \cite{toma_clinical_2023}, MedQA \cite{jin2021disease} & MMLU \cite{mmlu}, MedMCQA, MedQA, PubMedQA, USMLE Sample Exam \cite{usmle} \\

MedAlpaca \cite{han_medalpaca_2023} & 2023 & LLaMA & Medical Meadow dataset (introduced in paper). & USMLE assessment \\

PMC-LLaMA \cite{wu_pmc-llama_2023} & 2023 & LLaMA with IFT & MedC-K (introduced in paper): based on S2ORC \cite{Lo2020S2ORCTS} with emphasis on biomedical papers, 30000 textbooks. MedC-I (introduced, for IFT): based on MedAlpaca and ChatDoctor datasets, USMLE, PubMedQA, MedMCQA, UMLS medical knowledge graphs & MCQA: PubMedQA, MedMCQA, USMLE \\

LLaVA-Med \cite{li2024llava} & 2023 & fine-tuned LLaVA & Based on PMC-15M \cite{pmc15m} & Visual QA: VQA-RAD, SLAKE \cite{slake}, PathVQA \\

BioMistral \cite{biomistral} & 2024 & Mistral & PubMed Central Open Access & MMLU, MedQA, MedMCQA, PubMedQA \\

CheXagent \cite{CheXagent} & 2024 & custom, multimodal & CheXinstruct (introduced in paper), MIMIC-CXR \cite{mimiccxr}, PadChest \cite{PadChest}, BIMCV-COVID-19 \cite{vaya2020bimcv} & MIMIC-CXR, CheXpert \cite{irvin2019chexpert}, SIIM \cite{siimacr_pneumothorax}, RSNA \cite{rnsa}, OpenI \cite{demner2012design}, SLAKE \\


\bottomrule
    \end{tabular}

    \label{tab:medical_language_models_booktabs}
\end{adjustwidth}
\end{table}

Domain-specific \color{black} language models \color{black} are models tailored for a narrow field or topic, enhancing their ability to produce more accurate and relevant responses within that specific area. Generally, there are two common methods to create a domain-specific model: one approach is to pre-train a model using a set of domain-specific documents, such as medical papers; another is to take a generically trained model and fine-tune or adapt it to the target domain.

The medical field, rich in unstructured textual data from sources such as Electronic Health Records (EHR) and other medical documentation, presents an ideal research domain for language models to address a wide array of challenges. These applications range from tasks previously tackled by NLP techniques, such as clinical acronym disambiguation, to those unattainable a decade ago. For example, \color{black} medical conversational agents \color{black} that can assist both patients and healthcare professionals are now emerging. In response to these advancements, the research community has been actively developing language models specifically designed for medical applications. The evolution of these models, from pre-training and fine-tuning strategies to creating open models capable of zero-shot learning, highlights the growing synergy between artificial intelligence and healthcare \cite{zhang_alpacareinstruction-tuned_2023,li_chatdoctor_2023,abbasian_conversational_2023,toma_clinical_2023}.

Early medical language models were predominantly based on BERT and followed the common trend of the pre-training and fine-tuning paradigm. ClinicalBERT \cite{huang2020clinicalbert}, pre-trained on the MIMIC-III \cite{johnson2016mimiciii} dataset and fine-tuned to predict hospital readmissions, was one of the pioneering models in this domain. Another notable model, BioBERT, was pre-trained on the PubMed library and fine-tuned for tasks such as named entity recognition (NER), relation extraction (RE), and medical question answering (QA) \cite{Lee_2019}. The proven efficacy of pre-training language models on biomedical corpora and fine-tuning them for specific downstream tasks, like clinical concept extraction or measuring semantic text similarity (STS), led to the development of many other BERT-based models, including BiomedBERT \cite{chakraborty-etal-2020-biomedbert}, PubMedBERT \cite{Gu_2021}, BEHRT \cite{li2020behrt}, and GatorTron \cite{yang_gatortron_2022}.


The debut of BioGPT has marked another milestone in biomedical NLP. This \color{black} decoder-based model, \color{black} based on GPT-2 and trained on millions of PubMed abstracts, has demonstrated \color{black} proficiency \color{black} in specialized tasks, including text classification, and biomedical text generation \cite{luo_biogpt_2022}.

AlpaCare, a medical instruction-tuned LLM trained on more than fifty thousand instruction-response pairs generated artificially using GPT-4, represents a notable example from the current generation of medical LLMs that can be utilized in the medical domain without the need for fine-tuning on downstream tasks \cite{zhang_alpacareinstruction-tuned_2023}. Clinical Camel \cite{toma_clinical_2023}, another medical LLM fine-tuned from the Llama-2 model, excels in various medical tasks, ranging from clinical note creation to medical triaging. Despite its limitations, such as the potential for generating misleading content and the need for continuous updates, it marked a remarkable advancement in medical LLMs. The issue of needing continuous updates was addressed by the ChatDoctor \cite{li_chatdoctor_2023} model. Based on the Llama family and fine-tuned with real-world patient-doctor conversations, this model, \color{black}as part of an integrated framework, \color{black} can retrieve information from external sources, a crucial capability in the medical field, particularly for addressing emerging diseases.

Recent advancements in the field, particularly the development of multimodal models that work with both text and images, have substantially impacted the medical domain as well. The LLaVA-Med model \cite{li2024llava} can answer open-ended questions about biomedical images, while the CheXagent model \cite{CheXagent}, designed specifically for chest X-ray interpretation, exemplifies the successful transition from general medical models to specialist models capable of tackling problems in narrow medical fields.

Table~\ref{tab:medical_language_models_booktabs} provides a comprehensive, though not exhaustive, summary of existing medical language models. It highlights current trends in the development of medical models and the evolution of training and experimental datasets utilized. \color{black}The experimental datasets listed in the table offer valuable references for readers interested in evaluating performance on specific tasks within the medical domain.\color{black}

%


\section{Medical Applications of Language Models}

Medical applications of language models can be defined as the intersection of two sets: tasks that language models can accomplish and potential healthcare needs where predominantly textual language models can add value. Although both sets are finite, the cardinality of the first set is arguably smaller given the vast scope of the medical domain. Thus, our approach to categorizing medical applications is based on language models' tasks of various granularity, which, in turn, have a large area of intersection with NLP tasks in general. We focus on practically significant tasks from an application perspective, as well as notable macro applications that encompass multiple NLP tasks, deliberately excluding specific fine-grained NLP tasks such as coreference resolution and dependency parsing.

Table~\ref{tab:apps} provides an overview of selected applications of language models in the medical domain, focusing on clinical practice and related research. Subsequent subsections delve into each task, highlighting key applications, notable datasets, and exemplary implementations. Within the scope of this study, we intentionally exclude applications in medical education and highly specialized areas of biomedical research. Readers interested in a wider range of applications of language models in medicine may refer to the paper by Thirunavukarasu et al. \cite{Thirunavukarasu2023}.

\begin{table}[h!]
\footnotesize
\begin{adjustwidth}{-2.25in}{0in}
    \caption{Summary of Major Applications of Language Models in Medical Domain}
    \centering
    \begin{tabular}{p{3cm} p{7.5cm} p{4cm} >{\raggedright\arraybackslash}p{3cm}}
        \toprule
        \textbf{LLM Application} & \textbf{Examples of Medical Applications} & \textbf{Notable Datasets} & \textbf{Notable Solutions} \\
        \midrule
        Text Generation & Medical Report Generation \cite{WANG2023100033, chen2024diallama}, Clinical Note Generation \cite{brake2024comparing}, Generating Summaries For Laypersons \cite{eppler_bridging_2023}, Generating Summaries for Patient-Provider Dialogues\cite{nair2023generating}, Generating Textual Descriptions From Graph Models \cite{phatak2024narratingcausalgraphslarge} & CTRG-Chest-548K, CTRG-Brain-263K \cite{TANG2024121442}, IU-Xray \cite{demner2016preparing}, MIMIC-CXR \cite{johnson2019mimic}, CheXpert \cite{CheXpert} & Dia-LLaMA \cite{chen2024diallama}. Talk2Care \cite{talk2care}, MEDSUM-ENT \cite{nair2023generating}  \\
        \hline

        Token Classification & Clinical Acronym Disambiguation \cite{giacomini_clinical_2023, liu_exploring_2024}, Eponyms Disambiguation \cite{hagglund_classifiers_2023} & CASI \cite{CASI}, NLM-WSD \cite{weeber2001developing} & SciBERT, BioBERT, ClinicalBERT \cite{10.1007/978-3-031-36402-0_19} \\
        \hline
        
        Sequence Classification & Phenotyping \cite{groza2023evaluation}, Medical Coding \cite{lopez-garcia_explainable_2023}, Modeling Patient Timeline \cite{kraljevic_foresight_2023, shoham2023cpllm, dus_automated_2023}, Social Media Monitoring \cite{fisher} & Suicide Watch \cite{komati2019suicide}, CSSRS \cite{CSSRS}, MIMIC-III \cite{johnson2016mimiciii}, MIMIC-IV \cite{johnson2023mimiciv}, eICU-CRD \cite{eICUcrd}, HPO-GS \cite{hpogs}, BIOC-GS \cite{Weissenbacher2023}, CAMS \cite{garg-etal-2022-cams}, Wellness-Reddit\cite{liyanage}, Mental Disturbance\cite{Mentalrisk} & Foresight\cite{kraljevic_foresight_2023}, Bio\_ClinicalBERT\cite{xie_longterm_2023}, BioMedLM \cite{shoham2023cpllm} \\
        \hline

        Question Answering and Information Extraction & Querying Data from Electronic Health Records \cite{goel2023llms}, Extracting Information from Clinical Narrative Reports \cite{bhagat_large_2024}, Extracting Information From Medical Articles \cite{cao2023llm} & CASI \cite{CASI}, n2c2 \cite{n2c2}, i2b2 \cite{i2b2}, PubMedQA \cite{PubMedQA}, MedMCQA\cite{pal2022medmcqa}, emrQA \cite{pampari2018emrqa}, BIOASQ \cite{tsatsaronis2015overview} & quEHRy \cite{soni_quehry_2023}, BiomedBERT \cite{chakraborty-etal-2020-biomedbert}, PubMedBERT \cite{Gu_2021}, BioGPT \cite{luo_biogpt_2022}, Llava-med \cite{li2024llava}  \\
        \hline

        Summarization and Paraphrasing & Summarizing Clinical Study Reports \cite{10.1093/jamiaopen/ooae043}, Summarizing Patient-Provider Dialogues \cite{mishra-etal-2023-llm}, Simplification of Medical Texts \cite{devaraj-etal-2021-paragraph, artsi_large_2024}, Simplification of Radiology Reports \cite{Jeblick2024}, Improving Biomedical Text Readability \cite{swanson_biomedical_2024}, Anonymization of Medical Documents \cite{Wiest2024.06.11.24308355} & PLS-Cochrane Reviews \cite{devaraj-etal-2021-paragraph}, CELLS \cite{GUO2024104580}, Pfizer Clinical Trial Data \cite{10.1093/jamiaopen/ooae043}, MultiCohrane \cite{joseph2023multilingualsimplificationmedicaltexts} & fine-tuned BART \cite{devaraj-etal-2021-paragraph}, RALL \cite{GUO2024104580}, fine-tuned Llama \cite{10.1093/jamiaopen/ooae043,Wiest2024.06.11.24308355 } \\
        
        \hline
        
        Conversation & Mental Health Bots \cite{saha_towards_2022, yang_towards_2023, kang_domain-specific_2023}, Medical Chatbots and Health Assistants \cite{abbasian_conversational_2023, cung_performance_2024, nov_putting_2023}, Triaging \cite{Levine2023.01.30.23285067}, Differential Diagnosis \cite{kim2024human} & MotiVAte \cite{MotiVAte}, Depression\_Reddit \cite{dreddit}, CLPsych \cite{zirikly-etal-2019-clpsych}, Dreaddit \cite{dreaddit}, Clinical Vignettes\cite{Levine2023.01.30.23285067} & MedPaLM, DRG-LLaMA \cite{zhou2024interpretable}, openCHA \cite{abbasian_conversational_2023} \\

        \bottomrule
    \end{tabular}

\label{tab:apps}
\end{adjustwidth}
\end{table}

\subsection{Text Generation}
The task of text generation in the medical domain involves creating contextually accurate and relevant medical texts based on a sequence of prior tokens and specific context. This task may include generating clinical notes, patient reports, \color{black} or treatment plans\color{black}. The primary challenge is ensuring that the generated text is precise, medically accurate, and adheres to relevant privacy and ethical standards. 

\color{black} To achieve this, text generation models rely on a probabilistic framework\color{black}. The probability of generating the next token $x_t$, given a sequence of previous tokens $x_1, x_2, \ldots, x_{t-1}$ and additional context $c$, can be defined as:
\[
P(x_t \mid x_1, x_2, \ldots, x_{t-1}; c)
\]

In this formulation, \color{black}context $c$ \color{black}  represents various factors depending on the nature of the task. For instance, when generating patient reports, the context would include the available information about the patient, such as vital statistics, previous diagnoses and treatments. 

Decoder-based language models inherently generate text by predicting the probability of the next token based on preceding tokens and contextual information. The methods for providing context vary. The simplest approach is using prompts given directly to the language model. However, because the knowledge within language models is static, integrating external knowledge sources can be advantageous. This integration is often accomplished using variations of Retrieval-Augmented Generation (RAG) techniques \cite{10544639, jeong2024improving}.

Advancements in transformer-based models have led to innovative approaches in generating clinical documentation from patient-provider interactions. A study by Brake and Schaaf \cite{brake2024comparingmodeldesignsclinical} compares two model designs for generating clinical notes from doctor-patient conversations using the \color{black} encoder-decoder-based PEGASUS-X model \cite{pegasusx}\color{black}. The first design, GENMOD, generates the entire note in one step, while the second design, SPECMOD, generates each section independently. The study aims to evaluate the consistency of the generated notes in terms of age, gender, body part, and coherence. Evaluations were performed using ROUGE\cite{lin2004rouge} and Factuality\cite{factuality1} metrics, human reviewers, and the Llama2 LLM. Results indicate that GENMOD improves consistency in age, gender, and body part references, while SPECMOD may have advantages in coherence depending on the interpretation. The study uses a proprietary dataset with 10,859 doctor-patient conversations for training and testing \cite{brake2024comparingmodeldesignsclinical}. Nair et al. present MEDSUM-ENT\cite{nair2023generating}, a multi-stage approach to generating medically accurate summaries from patient-provider dialogues. Using GPT-3 as the backbone, the approach first extracts medical entities and their affirmations from conversations and then constructs summaries based on these extractions through prompt chaining. The model leverages ICL and dynamic example selection to improve entity extraction and summarization. The dataset used for evaluation consists of 100 de-identified clinical encounters from a telehealth platform. MEDSUM-ENT demonstrates improved clinical accuracy and coherence in summaries compared to a zero-shot, single-prompt baseline, as evidenced by both qualitative physician assessments and quantitative metrics designed to capture medical correctness \cite{nair2023generating}.


\color{black}The integration of LLMs with visual models facilitates the automatic generation of medical imaging reports by aligning visual features from models like Swin Transformer \cite{liu2021swintransformerhierarchicalvision} with the LLM's word embedding space, enabling effective multi-modal understanding\color{black}. Chen et al. developed Dia-LLaMA \cite{chen2024diallama}, a framework that utilizes the LLaMA2-7B model combined with a pre-trained ViT3D \cite{dosovitskiy2021imageworth16x16words} to manage high-dimensional CT data. It features a disease prototype memory bank and a disease-aware attention module to counteract the imbalance in disease occurrence. The framework, tested on the CTRG-Chest-548K dataset \cite{TANG2024121442}, outperformed other methods in various natural language generation metrics presented in the study. Another approach, R2GenGPT \cite{WANG2023100033}, enables radiology report generation by utilizing a visual alignment module that aligns visual features from chest X-ray images with the word embedding space of LLMs, thereby enhancing the capability of static LLMs to process visual data. It explores three alignment strategies: shallow, deep, and delta, each varying in trainable parameters. Evaluated on the IU-Xray and MIMIC-CXR datasets, R2GenGPT achieved impressive results in model efficiency and clinical metrics, leveraging the Swin Transformer\cite{liu2021swintransformerhierarchicalvision} and Llama2-7B model for enhanced integration.

Evaluating commercial models for clinical tasks inevitably draws the interest of the scientific community. Ali et al. evaluated the use of ChatGPT for generating patient clinic letters, focusing on its readability, factual correctness, and human-like quality by testing the model with shorthand instructions simulating clinical input for creating letters addressing skin cancer scenarios. The study involved 38 hypothetical clinical scenarios, including basal cell carcinoma, squamous cell carcinoma, and malignant melanoma. Readability was assessed using the online tool Readable\cite{readable}, targeting a sixth-grade reading level. Two independent clinicians evaluated the letters' correctness and human-like quality using a Likert scale. The study found that ChatGPT-generated letters scored highly in correctness and human-like quality \cite{ali_using_2023}.

Overall, while we observe the significant potential of text-only LLMs in text generation tasks, the emergence of multimodal models will inevitably bear fruit in this class of tasks in the medical domain. For example, fully multimodal models will be able to accurately generate clinical documentation based on patient-provider verbal dialogues and summarize them based on the end user's needs, while generating medical imaging reports will become less complicated and more accurate when done by a single multimodal model.

\subsection{Token Classification}

Token classification tasks in the medical domain involve labeling individual words or phrases within a text with specific medical annotations, such as identifying and disambiguating medical conditions, medications, dosages, and symptoms from clinical text.

\color{black} Given a sequence of tokens \( X = (x_1, x_2, \ldots, x_n) \) and context $c$, the task is to assign a label \( y_i \in \mathcal{Y}  \) to each token \( x_i \), where \( \mathcal{Y}  \) is the set of possible categories. This can be expressed within a probabilistic framework as:

\[
y_i = \arg\max_{k \in \mathcal{Y} } P(k \mid x_i, c),
\]

where \( P(k \mid x_i, c) \) is the probability of token \( x_i \) belonging to class \( k \). In this formulation, context $c$ may refer to the surrounding tokens or encompass information outside of the sequence \(X\), such as external vocabulary. 

Encoder-based models, such as BERT, typically excel in this category of tasks. However, they require fine-tuning on a labeled dataset to adapt to specific tasks, and fine-tuning may need to be reiterated if new information or data becomes available. In contrast, models with a decoder component, can use prompts to generate text annotated with labels, thereby directly producing labeled sequences without additional task-specific fine-tuning.
\color{black}

The widespread use of medical abbreviations and acronyms often leads to misunderstandings, necessitating accurate disambiguation of these terms to safeguard against misinterpretations that could jeopardize patient care \cite{parry_abbreviation_2023}. The process of mapping the short forms of medical terms to their full expressions is referred to as clinical acronym disambiguation. Wang and Khanna \cite{10.1007/978-3-031-36402-0_19} evaluated the performance of various clinical BERT-based language models on the Clinical Acronym Sense Inventory (CASI) dataset and found that ClinicalBert addresses the task effectively, achieving an F1 score of 0.915. In contrast, Sivarajkumar et al. \cite{sivarajkumar_empirical_2023} assessed the capabilities of generative LLMs, including GPT3.5, Bard (Gemini), and Llama2, in acronym disambiguation using the same dataset. Their study revealed that these models perform well in acronym disambiguation without fine-tuning, with GPT3.5 achieving the highest accuracy of 0.96.

While the problem of clinical acronym disambiguation might seem largely addressed, several challenges persist. For instance, Kugic et al.\cite{giacomini_clinical_2023} conducted a study on clinical acronym disambiguation in German using ChatGPT and Bing Chat (Copilot), achieving an F1 score of 0.679, which highlights the need for improvement. Another concern involves the datasets used in experiments. There is a possibility that LLMs may memorize specific terms, which could misrepresent their true disambiguation capabilities \cite{bordt2024elephants}\cite{ranaldi2023precog}. This issue necessitates further investigation to ensure the reliability of LLMs in clinical abbreviation disambiguation tasks with realistic datasets.

\subsection{Sequence Classification}

\color{black}
Sequence classification tasks in the medical domain involve assigning a label or multiple labels to an \textit{entire sequence} of text, rather than to individual tokens. These tasks may include classifying clinical documents or patient notes into categories such as diagnosis, treatment recommendation, or urgency level.

Most sequence classification tasks fall into the category of single-label classification, which can be formulated as follows: given a sequence of tokens \( X = (x_1, x_2, \ldots, x_n) \), the task is to assign a label \( y \in \mathcal{Y} \), where \( \mathcal{Y} \) is the set of possible categories:

\[
y = \arg\max_{k \in \mathcal{Y} } P(k \mid X),
\]

where \( P(k \mid X) \) represents the probability of the sequence \( X \) belonging to class \( k \).

Encoder-based models are well-suited for sequence classification tasks. For example, BERT can be fine-tuned for classification tasks by adding a fully connected layer on top of the embedding of the [CLS] token, which serves as a contextual representation of the entire sequence. Alternatively, pooling functions such as average pooling or max pooling can be applied to the token embeddings of the sequence to aggregate information across the entire input. Similar to token classification tasks, sequence classification often requires re-fine-tuning when new data or updated label distributions become available.

Generative language models, which incorporate a decoder component, can be either simply prompted or fine-tuned to predict class labels as part of the sequence. Fine-tuning approach typically involves training the model on inputs where a special token or delimiter demarcates the end of the input sequence and the beginning of the output label. In some cases, integrating a linear layer after the final token output in generative models can refine the logits corresponding to class predictions, enhancing the classification boundaries produced by the model's generative framework. However, this approach introduces additional complexity and is less commonly used.
\color{black}

The array of tasks that fall under sequence classification is vast. The following subsections detail a few of the most prominent applications.

\subsubsection{Suicidal Behavior Prediction}

Suicidal behavior prediction tasks predominantly focus on analyzing individuals' social media activities. Dus and Nefedov \cite{dus_automated_2023} proposed an automated tool for identifying potential self-harm indications in social media posts, \color{black} framing the task as a binary classification problem. The set of possible categories \( \mathcal{Y}  \) includes two elements: \( 1 \), indicating the presence of suicidal behavior, and \( 0 \), indicating its absence. The input, a sequence of tokens \( X \), is derived from the text of social media posts. The model estimates \( P(y = 1 \mid X) \), the probability that the input suggests suicidal behavior, by leveraging a fine-tuned ELECTRA model. Training data included samples from Kaggle's ``Suicide Watch'' dataset \cite{komati2019suicide}, supplemented with additional social media sources. The proposed method achieved an accuracy of \( 0.93 \) and an F1 score of \( 0.93 \) \cite{dus_automated_2023}.
\color{black}

Beyond mere social media post analysis, Levkovich et al. \cite{levkovich_suicide_2023} assessed ChatGPT-3.5 and ChatGPT-4’s ability to evaluate suicide risk based on perceived burdensomeness and thwarted belongingness. By comparing ChatGPT’s assessments to those made by mental health professionals using vignettes, they discovered that ChatGPT-4’s evaluations were in close alignment with professional judgments. In contrast, ChatGPT-3.5 tended to underestimate suicide risk, underscoring the limitations of these models in this specific area.

In summary, while treating suicidal behavior identification as a straightforward classification task on social media posts can lead to impressive scores using standard classification metrics, the practical and ethical implications of such approaches, including potential breaches of autonomy and principles of non-maleficence, are debatable \cite{fairness}. Well-structured vignette studies on the effectiveness of LLMs and other models can further advance research in this area. Additionally, exploring the potential of human-AI collaboration represents another promising research direction in this field.

\subsubsection{Modeling Patient Timeline}

The task of modeling patient timelines is multifaceted, involving forecasting future medical events, understanding patient trajectories, and predicting medical outcomes. This endeavor employs deep learning, transformers, and generative models to analyze data from various medical records, both structured and unstructured.

Kraljevic et al. introduced Foresight \cite{kraljevic_foresight_2023}, a GPT-2-based pipeline developed for modeling biomedical concepts extracted from clinical narratives. This pipeline employs NER and linking tools to transform unstructured text into structured, coded concepts. Utilizing datasets from three hospitals, covering over 800,000 patients, Foresight showed promise in forecasting future medical events. Its effectiveness was manually validated by clinicians on synthetic patient timelines, highlighting its potential in real-world risk forecasting and clinical research.

Among different types of models, generative adversarial networks (GANs) have gained popularity, extending their applications beyond the initial domain of image generation. Shankar et al. proposed Clinical-GAN \cite{johnson2023mimiciv}, which merges Transformer and GAN methodologies to model patient timelines, focusing on predicting future medical events based on past diagnosis, procedure, and medication codes. Tested on the MIMIC-IV dataset, Clinical-GAN outperformed baseline methods in trajectory forecasting and sequential disease prediction \cite{shankar_clinical-gan_2023}. Another study \cite{kadri_towards_2023} employed GAN for predicting the length of stay in emergency departments. The learning process was done in multiple stages. Initially, an unsupervised training phase used a generator and discriminator to approximate the probability distribution and perform feature discovery and reconstruction. Discriminator was then fine-tuned to optimize its parameters for global optimum. A predictor layer, initially randomly initialized, was added and optimized during fine-tuning, enabling the model to map observations to their lengths of stay. The model was trained on data from the Pediatric Emergency Department in CHRU-Lille and proved the potential of GANs in this field.

\textit{Medical outcome prediction} can be seen as a subtask of modeling patient timeline and is often scoped to either predict mortality, outcomes of a specific disease, or risk of progression from one disease to another. A recent study by Shoham and Rappoport \cite{shoham2023cpllm} examined data related to chronic kidney disease, acute and unspecified renal failure, and adult respiratory failure from the MIMIC-IV and eICU-CRD datasets. Using this data, the team generated labeled datasets for disease diagnosis prediction based on patient histories. They introduced a method named Clinical Prediction with Large Language Models (CPLLM) by fine-tuning LLMs (Llama2 and BioMedLM) using medical-specific prompts to help the models understand complex medical concept relationships. Xie et al. \cite{xie_longterm_2023} used EHR analysis to predict epilepsy seizures, leveraging Bio\_ClinicalBERT, RoBERTa, and T5, achieving an F1 score of $0.88$ in outcome classification.

A notable approach to predict outcomes of COVID-19 patients was proposed by Henriksson et al. \cite{henriksson_multimodal_2023}. The authors created a model that combines structured data and unstructured clinical notes in a multimodal fashion, leveraging a clinical KB-BERT model for multimodal fine-tuning. Trained on data from six hospitals in Stockholm, Sweden, their model effectively predicted 30-day mortality, safe discharge, and readmission of COVID-19 patients in the emergency department, \color{black}as measured by AUC  \color{black} .

\subsubsection{Phenotyping and Medical Coding}
The phenotyping task primarily involves identifying phenotypic abnormalities from a patient's various medical records, which aids in the identification of rare diseases. There exists the Human Phenotype Ontology (HPO) project\footnote{\url{https://hpo.jax.org}} that systematically categorizes human phenotypes with detailed annotations. \color{black} The phenotyping task can be framed as a \textbf{multi-label classification task}, which extends the single-label classification task as follows. Given a sequence of tokens \( X = (x_1, x_2, \ldots, x_n) \), the task is to assign a subset of labels \( Y \subseteq \mathcal{Y}  \), where \( \mathcal{Y}  \) is the set of all possible labels.

For each label \( k \in \mathcal{Y}  \), the model predicts a probability \( P(k \mid X) \) representing the likelihood of the sequence \( X \) being associated with label \( k \). The set of labels \( Y \) is typically determined by applying a predefined threshold \( \theta \) to these probabilities:

\[
Y = \{ k \in \mathcal{Y}  \mid P(k \mid X) \geq \theta \}.
\]

Thus, in the phenotyping task, the sequence \( X \) represents a medical record of a patient, and \( \mathcal{Y}  \) denotes the set of all possible phenotype labels from the HPO. The objective is to identify a subset \( Y \subseteq \mathcal{Y}  \) of HPO labels associated with the medical record.
\color{black}

Traditionally, the task of phenotyping has relied on named entity recognition, where models similar to BERT have demonstrated proficiency. However, recent studies have started exploring in-context learning and zero-shot learning with contemporary LLMs, yielding promising results \cite{groza2023evaluation}.

Medical coding is another multi-label classification task that involves identifying a set of International Classification of Diseases (ICD) codes\footnote{\url{https://icd.who.int}} associated with a medical record. This task can be formulated similarly to phenotyping, with the key difference being that \( \mathcal{Y}  \) represents a set of ICD codes rather than HPO labels. Besides observing trends similar to those in the phenotyping subdomain, it is also noteworthy that there is a shift towards explainable medical coding, as highlighted in \cite{lopez-garcia_explainable_2023}.

\subsection{Question Answering and Information Extraction}

Question Answering (QA) task can be formulated as finding the answer \(A\) from a possible set of answers \(\mathcal{A}\), given a question \(q\) and a context \(c\) (often a document or set of documents containing information relevant to the question). This can be expressed as:
\[
A = \operatorname{argmax}_{a \in \mathcal{A}} P(a \mid q, c)
\]
where \(P(a \mid q, c)\) is the probability of \(a\) being the correct answer given the question \(q\) and the context \(c\).

Information Extraction (IE) task involves identifying specific pieces of information (entities, relationships, events) within documents. This can be described as a function \(f\) that maps a set of documents \(D\) to a set of structured attributes \(S\), which includes entities \(E\), relationships \(R\), and other attributes of interest:
\[
S = f(D) = \{E, R, \ldots\}
\]
Here, \(D\) is the input document, and \(S\) represents the structured output containing extracted elements.

Encoder-based models are well-suited for question answering tasks. They can be fine-tuned on specific QA datasets, where the input is a concatenation of the question and context (a paragraph or document containing the answer). The model is then trained to identify the span of text that answers the question, typically by adding a start and end token classifier to the output embeddings of the model. These classifiers predict the beginning and end positions of the answer in the text. On the other hand, generative models with a decoder component leverage their extensive pre-training on diverse data. By inputting a question (along with the document of interest when necessary) and following it with a prompt that encourages the model to generate an answer, these models can produce responses without needing explicit pointers to answer spans.

In the medical domain, IE and QA systems are instrumental for extracting data from electronic health records, such as medication lists and diagnostic details, essential for patient management and treatment planning. A notable example is quEHRy, a QA system designed to query EHRs using natural language. The primary goal of quEHRy is to provide precise and interpretable answers to clinicians' questions from structured EHR data \cite{soni_quehry_2023}. Beyond the successful applications of BERT-based models like BioBERT, BiomedBERT, and PubMedBERT for QA and IE, generative models also show proficiency. Agrawal et al. \cite{agrawal_large_2022} demonstrated the effectiveness of generative LLMs such as InstructGPT and GPT-3 in zero-shot and few-shot information extraction from clinical texts, \color{black}measured through accuracy and F1 score\color{black}. When tested on the re-annotated CASI dataset, these models showed considerable potential in tasks requiring structured outputs. Furthermore, Ge et al. \cite{ge_comparison_2023} compared the effectiveness of LLMs versus manual chart reviews for extracting data elements from EHRs, focusing specifically on hepatocellular carcinoma imaging reports. Using the GPT-3.5-turbo model, implemented as ``Versa Chat" within a secure UCSF \footnote{\url{https://www.ucsf.edu/}} environment to protect patient health information, the study analyzed 182 CT or MRI abdominal imaging reports from the Functional Assessment in Liver Transplantation study. It extracted six distinct data elements, including the maximum LI-RADS\footnote{\url{https://www.acr.org/Clinical-Resources/Reporting-and-Data-Systems/LI-RADS} } score, number of hepatocellular carcinoma lesions, and presence of macrovascular invasion. The performance was evaluated by calculating accuracy, precision, recall, and F1 scores, showing high overall accuracy (0.889) with variations depending on the complexity of the data elements.

\subsection{Summarization and Paraphrasing}

Paraphrasing involves rewriting a text \( T \) into a new form \( P \), ensuring that \( P \) maintains the same meaning as \( T \) but utilizes different vocabulary and potentially altered sentence structures. Summarization, on the other hand, entails generating a brief version of a text \( T \) that preserves its core information. Abstractive summarization can be considered a specific case of paraphrasing.

Encoder-based models are proficient at extractive summarization. They evaluate sentences within a text to determine their relevance and informativeness. By scoring each sentence, these models identify and concatenate the most important sentences to form a coherent summary. In contrast, abstractive summarization and paraphrasing typically employ decoder-based or sequence-to-sequence (encoder-decoder) models. These models are trained to understand the entire narrative or document and then recreate its essence in a different form.

Summarization and paraphrasing tools are used in the medical domain for managing extensive documentation and enhancing communication. Summarization helps healthcare professionals quickly grasp essential details from lengthy clinical notes, generate concise abstracts of medical research papers, and craft clear patient discharge summaries, thereby improving patient comprehension and adherence to medical advice. Paraphrasing makes complex information more accessible by translating medical jargon into simpler language for patient education. It also enhances the clarity and consistency of electronic health records, aiding healthcare providers in better understanding and utilizing the data effectively.

Summarization and paraphrasing in the medical domain are largely driven by advancements in these tasks in general. Devaraj et al. \cite{devaraj-etal-2021-paragraph} introduce a new dataset derived from the Cochrane Database of Systematic Reviews\footnote{\url{https://www.cochranelibrary.com/cdsr/about-cdsr}}, featuring pairs of technical abstracts and plain language summaries. They propose a novel metric based on encoder-based language models to better distinguish between technical and simplified texts. The study utilizes baseline encoder-decoder Transformer models for text simplification and introduces an innovative approach to penalize the generation of jargon terms. The code and data are publicly available for further research.

The paper titled ``Biomedical Text Readability After Hypernym Substitution with Fine-Tuned Large Language Models" investigates simplifying biomedical text using LLMs to enhance patient understanding. The authors fine-tuned three LLM variants to replace complex biomedical terms with their hypernyms. The models used include GPT-J-6b, SciFive T5 \cite{phan2021scifivetexttotexttransformermodel}, and an approach combining sequence-to-sequence and sciBERT \cite{beltagy2019scibertpretrainedlanguagemodel} models. The study processed 1,000 biomedical definitions from the Unified Medical Language System (UMLS) and evaluated readability improvements using metrics such as the Flesch-Kincaid Reading Ease and Grade Level, Automated Readability Index, and Gunning Fog Index. Results showed substantial readability improvements, with the GPT-J-6b model performing best in reducing sentence complexity \cite{swanson_biomedical_2024}.

Another interesting application of paraphrasing is the anonymization of medical documents, which is crucial for balancing ethical principles and research needs. Wiest et al.\cite{Wiest2024.06.11.24308355} present an approach to de-identify medical free text using LLMs. The authors benchmarked eight locally deployable LLMs, including Llama-3 8B, Llama-3 70B, Llama-2 7B, Llama-2 70B, and Mistral 7B, on a dataset of 100 clinical letters from a German hospital. They developed the LLM-Anonymizer pipeline, which achieved a success rate of 98.05\% in removing personal identifying information using Llama-3 70B. The tool is open-source, operates on local hardware, and does not require programming skills, making it accessible and practical for use in medical institutions. The study demonstrates the potential of LLMs to effectively de-identify medical texts, outperforming traditional NLP methods and providing a robust solution for privacy-preserving data sharing in healthcare.


Despite advancements in summarization and paraphrasing, some challenges persist, particularly in preserving factual accuracy and precision. Jeblick et al.\cite{Jeblick2024} explored the effectiveness of using ChatGPT  (version December 15th, 2022) to simplify radiology reports into language understandable by non-experts. A radiologist created three hypothetical radiology reports, which were then simplified by prompting ChatGPT. Fifteen radiologists evaluated the quality of these simplified reports based on criteria such as factual correctness, completeness, and potential harm to patients. The study used Likert scale analysis and inductive free-text categorization to assess the simplified reports. Overall, the radiologists found the simplified reports to be factually correct and complete, with minimal potential for harm. However, some issues were noted, including incorrect information, omissions of relevant medical data, and occasionally misleading or vague statements. These issues highlight the need for careful supervision by medical professionals when using language models to simplify complex medical texts. A recent study by Landman et al.\cite{10.1093/jamiaopen/ooae043} discusses a challenge organized by Pfizer to explore the use of LLMs for automating the summarization of safety tables in clinical study reports. Various teams employed GPT models with prompt engineering techniques to generate summary texts. The datasets included safety outputs from 72 reports from recent clinical studies, split into 70\% for training and 30\% for testing. The study concluded that while LLMs show promise in automating the summarization of clinical study report tables, human involvement and further research are necessary to optimize their application.

\subsection{Conversation}
\color{black} The task of conversation, or dialogue generation, can be formulated as follows. Given a dialogue history \( H = (h_1, h_2, \ldots, h_n) \), where each \( h_i \) represents an utterance in the conversation, the objective is to generate an appropriate response \( R \). This can be expressed as:

\[
R = \operatorname{argmax}_{r} P(r \mid H),
\]

Where \( P(r \mid H) \) represents the conditional probability of generating the response \( r \) given the dialogue history \( H \).
\color{black} 
Typically, pre-trained decoder-based large language models are fine-tuned using specialized datasets to develop their conversational capabilities. In the medical domain, conversational applications facilitate interactive communication with patients. For example, conversational AI can be deployed as virtual health assistants that provide initial consultations based on symptoms described by patients. These systems can ask relevant follow-up questions, assess symptoms, and offer preliminary advice or direct patients to seek professional care when necessary. Additionally, these conversational tools can be utilized for patient education, explaining complex medical conditions and treatments in simple language to enhance understanding and compliance. Another notable application is in mental health support, where conversational AI can offer coping strategies and basic support, thereby augmenting traditional therapy sessions.

\subsubsection{Chatbots and Health Assistants}

The proficiency of LLMs in generating coherent text and finding patterns in natural language makes them excellent candidates for Conversational Health Agents (CHAs) or chatbots. The impressive capabilities of systems like ChatGPT have sparked researchers' interest in evaluating them as out-of-the-box medical chatbots. These chatbots are capable of holding conversations on medical topics and providing valid, science-based responses, akin to human doctors. 

Cung et al.\cite{cung_performance_2024} assessed the performance of three commercial systems - ChatGPT, Bing (Copilot), and Bard (Gemini) in the context of skeletal biology and disorders. The study involved posing 30 questions across three categories, with the responses graded for accuracy by four reviewers. While ChatGPT 4.0 had the highest overall median score, the study revealed that the quality and relevance of responses from all three chatbots varied widely, presenting issues such as inconsistency and failure to account for patient demographics. Another study explored using ChatGPT for patient-provider communication. A survey of 430 participants found that ChatGPT responses were often indistinguishable from those of healthcare providers, indicating a level of trust in chatbots for answering lower-risk health questions \cite{nov_putting_2023}.

Despite the success of chatbots in general and low-risk medical interactions, studies suggest that chatbots are not yet suitable for high-risk subdomains. For instance, a study focusing on resuscitation advice provided by Bing (Copilot) and Bard (Gemini) chatbots revealed that the responses frequently lacked guideline-consistent instructions and occasionally contained potentially harmful advice. Only a small fraction of responses from Bing (9.5\%) and Bard (11.4\%) completely met the checklist criteria ($P > .05$), underscoring the current limitations of LLM-based chatbots in critical healthcare scenarios \cite{birkun_large_2023}. 

Another research direction in the field of medical chatbots is the implementation of conversational agents specifically designed for the medical domain. Abbasian et al.\cite{abbasian_conversational_2023} proposed a complex LLM-based multimodal framework for CHAs, concentrating on critical thinking, knowledge acquisition, and multi-step problem-solving. This framework aims to enable CHAs to provide personalized healthcare responses and handle intricate tasks such as stress level estimation.

Domain-specific LLMs, such as the ChatDoctor model \cite{li_chatdoctor_2023}, can also serve as chatbots. This model integrates a self-directed information retrieval mechanism, allowing it to access up-to-date information from online and curated offline medical databases. Evaluated using BERTScore\cite{bertscore}, ChatDoctor exhibited a higher F1 score compared to ChatGPT-3.5, demonstrating the effectiveness of smaller domain-specific models as alternatives to large commercial solutions. 

Overall, chatbots show promise, particularly in low-risk consultation areas. However, concerns such as potential confabulations, lack of explainability, and biases highlight challenges in their application in real-case scenarios \cite{10.1093/cid/ciad633}. Additionally, the absence of a robust, comprehensive, and universally accepted evaluation metric for chatbots is notable. Human evaluation lacks scalability, and similarity metrics like BERTScore may overlook critical factual inaccuracies.

\subsubsection{Mental Health Bots}
The idea of using a machine as a personal psychologist dates back to at least the 1960s when Weizenbaum proposed a simple rule-based system called ELIZA \cite{eliza}. Contemporary advancements in mental health chatbots are largely driven by LLMs. Yang et al.\cite{yang_towards_2023} investigated the capabilities of current LLMs in automated mental health analysis. Their study involved evaluating LLMs across diverse datasets for tasks such as emotional reasoning and detecting mental health conditions, employing various similarity metrics including BLEU, ROUGE family, BERTScore derivatives, BART-score \cite{bartscore}, and human assessments. They discovered that while ChatGPT displays robust in-context learning abilities, it still encounters challenges in emotion-related tasks and requires careful prompt engineering to improve performance to enhance its performance.

Saha et al. \cite{saha_towards_2022} introduced a Virtual Assistant for supporting individuals with Major Depressive Disorder, using a dataset called MotiVAte. Their system, based on modified GPT-2 model and reinforced learning, shows promising results in generating empathetic and motivational responses, as evidenced by both automated evaluations based on text similarity and human evaluations based on fluency, adaptability, and degree of motivation. Sharma et al.\cite{sharma-etal-2023-cognitive} introduced a dataset for training a GPT-3-based model for generating reframes with controlled linguistic attributes. Deployed on the Mental Health America website\footnote{\url{https://mhanational.org/}}, this allowed for a randomized field study to gather findings on human preferences. Another team explored the fine-tuning of open-source LLMs on psychotherapy assistant instructions, using a dataset from Alexander Street Press\footnote{\url{https://alexanderstreet.com/}} therapy and counseling sessions. Their results indicated that LLMs fine-tuned on domain-specific instructions surpassed their non-fine-tuned counterparts in psychotherapy tasks, underscoring the significance of professional and context-specific training for these models \cite{kang_domain-specific_2023}.

Promising outcomes have been observed through collaborations between humans and AI. A recent study\cite{sharma2022humanai} conducted a randomized controlled trial involving human peer supporters, demonstrating that an AI-in-the-loop agent led to a 19.60\% increase in conversational empathy in interactions between individuals seeking mental health support and support specialists. This was achieved by providing suggestions for response improvements to peer supporters. This research reveals that human-AI collaboration is a crucial area for potential exploration, particularly in the medical domain.

The evolving reasoning capabilities of LLMs have sparked interest in their use for disease diagnostics. Levine et al.\cite{Levine2023.01.30.23285067} conducted experiments with the GPT-3 model to assess its diagnostic and triage accuracy. Their results indicate that GPT-3's diagnostic accuracy is comparable to that of physicians but lags in triage accuracy. GPT-3 correctly identified the diagnosis in its top three choices for 88\% of the cases, surpassing non-experts (54\%) but slightly underperforming compared to professional physicians (96\%). In triage performance, GPT-3 achieved an accuracy of 70\%, on par with non-experts (74\%) but significantly lower than physicians (91\%). Despite GPT-3's notable performance, the study raises ethical concerns, particularly regarding the model's potential to perpetuate existing data biases, exhibiting racial and gender biases and occasionally producing misleading or incorrect information.

A recent study by Liu et al. \cite{liu2023pharmacygpt} introduced a framework named PharmacyGPT, which leverages the current GPT family models to emulate the role of clinical pharmacists. This research utilized real data from the ICU at the University of North Carolina Chapel Hill (UNC) Hospital. PharmacyGPT was applied to tackle various challenges in the realm of pharmacy, encompassing patient outcome studies, AI-based medication prescription generation, and interpretable patient clustering analysis. The study revealed that the GPT-4 model, when provided with dynamic context and similar samples, attained the highest accuracy among all models tested. However, the precision and recall scores were not notably high across the approaches. This outcome may be caused by the binary nature of mortality prediction, a significant imbalance in the dataset, and the complex, individualized nature of ICU pharmacy regimens. The research highlights the need for custom evaluation metrics to assess the performance of AI-generated medication plans, enhancing understanding of the models' strengths and limitations.

\section{Discussion}
This section explores the challenges and opportunities arising from the integration of large language models in healthcare.

\subsection{Evaluation Challenges}
\color{black}
The evaluation of language models in the medical domain is a multifaceted challenge. One major issue arises from the technical complexity of assessing model performance on tasks with minimal human supervision. For instance, while classification tasks benefit from well-established metrics such as accuracy, precision, recall, and F-measure, evaluating models on more complex tasks, such as medical conversations, remains technically challenging. These tasks often require human assessment or intricate evaluation frameworks \cite{cung_performance_2024, nov_putting_2023, dynaeval, ghazarian2022deam}.

Another challenge arises from unique domain-specific issues, where standard model evaluation from a purely technical perspective is often insufficient. Abbasian et al. \cite{abbasian2024foundation} categorize evaluation metrics into \textit{intrinsic} metrics, such as Perplexity, which measure internal language proficiency and coherence, and \textit{extrinsic} metrics, which assess real-world impact and the model's ability to meet human-centric expectations. Within extrinsic metrics, the authors identify general metrics applicable to all tasks, such as trustfulness, bias, and toxicity, as well as domain-specific metrics, such as up-to-dateness and empathy. While the intrinsic/extrinsic categorization provides a useful theoretical framework for understanding evaluation metrics, in practice, metrics are often organized based on their functional role in model evaluation. Table~\ref{tab:evaluation_metrics} presents a task-oriented categorization, where metrics are grouped by their evaluation purpose. 

\begin{table}[h!]
\footnotesize
\begin{adjustwidth}{-2.25in}{0in}
    \caption{Evaluation Metrics for Major Tasks in Medical Domain using Language Models}
    \centering
    \begin{tabular}{p{7cm} p{3.5cm} p{3.5cm} p{3cm}}
        \toprule
        \textbf{Tasks} & \textbf{Task-Specific Metrics} & \textbf{Extrinsic Generic Metrics and Benchmarks} & \textbf{Extrinsic Domain-Specific Metrics and Benchmarks} \\
        \midrule

        \textbf{Classification and Information Extraction:} Clinical Acronym Disambiguation, Eponyms Disambiguation, Phenotyping, Medical Coding, Modeling Patient Timeline, Social Media Monitoring, Querying Data from Electronic Health Records, Extracting Information from Clinical Narrative Reports, Extracting Information From Medical Articles & Accuracy, Precision, Recall, F-Score, AUC & Necessary Only When Using Generative Models: Metrics for evaluating LLM performance in general (e.g., Academic Benchmarks, Factuality, Complex Reasoning) \cite{devaraj2022evaluating, geminiteam2024}. & Privacy \cite{lukas2023analyzing} (when patient data is involved) \\ 
        \hline

        \textbf{Generation and Summarization:} Medical Report Generation, Clinical Note Generation, Generating Summaries For Laypersons, Generating Summaries for Patient-Provider Dialogues, Generating Textual Descriptions From Graph Models, Summarizing Clinical Study Reports, Summarizing Patient-Provider Dialogues, Simplification of Medical Texts, Simplification of Radiology Reports, Improving Biomedical Text Readability  & BLEU, ROUGE, METEOR, Perplexity \cite{jm3}, BERTScore & Metrics for evaluating LLM performance in general (e.g., Academic Benchmarks, Factuality, Complex Reasoning). & Reliability \cite{abbasian2024foundation}, Up-to-dateness \cite{petroni2019language}, Privacy \\ 
        \hline

        \textbf{System-Patient Conversation:} Mental Health Bots, Medical Chatbots and Health Assistants & Match Rate, Dialogue Accuracy, Average Request Turn, Complex Chatbot-Specific Metrics (e.g., DEAM \cite{ghazarian2022deam}, DynaEval \cite{dynaeval}) & Metrics for evaluating LLM performance in general (e.g., Academic Benchmarks, Factuality, Complex Reasoning), Safety and Bias \cite{dhamala2021bold, parrish2022bbq, abbasian2024foundation, gehman2020realtoxicityprompts, lin2022truthfulqa}.  & Reliability, Up-to-dateness, Privacy, Empathy \cite{abbasian2024foundation}  \\
        
        \bottomrule
    \end{tabular}
\label{tab:evaluation_metrics}
\end{adjustwidth}
\end{table}

Unlike task-specific metrics, which are widely adopted and often have clear mathematical formulations, extrinsic metrics involving evaluation of LLMs are significantly more technically complex due to their reliance on specialized frameworks, human alignment, or external models. For example, DynaEval \cite{dynaeval} is a framework-based metric that employs a graph convolutional network to evaluate model performance.  Benchmarks designed for specific goals, such as TruthfulQA \cite{lin2022truthfulqa}, RealToxicityPrompts \cite{gehman2020realtoxicityprompts}, and BBQ \cite{parrish2022bbq}, may fail to generalize well to multilingual tasks or domain-specific applications due to their reliance on carefully crafted datasets tailored to specific scenarios. Furthermore, some extrinsic metrics rely on external models for evaluation \cite{ghazarian2022deam}, introducing stochasticity into the process due to the non-deterministic nature of model outputs.

Moreover, the probabilistic nature of LLMs makes it challenging to evaluate their performance with respect to domain-specific metrics, especially for unseen input combinations or highly specialized, out-of-distribution data.

Notably, the demand for domain-specific extrinsic metrics grows as models are opened to a broader audience performing more complex tasks. In the healthcare sector, domain-specific evaluation is particularly critical due to its unique ethical and regulatory requirements. Hond et al. \cite{de2024text} highlight the importance of complementing general and domain-specific evaluation with \textit{clinical impact validation}, a process that assesses outcomes such as improved health results, higher patient satisfaction, or reduced administrative burdens. However, to the best of our knowledge, no existing frameworks can automatically perform clinical impact validation with reasonable accuracy.

\color{black}

\subsection{Ethical Issues}
\color{black}

The application of language models in clinical settings raises substantial ethical concerns. To address these, foundational principles such as respect of persons, beneficence, and justice, as outlined in the Belmont Report \cite{belmont1979}, provide a guiding framework. Solomonides et al. \cite{Solomonides} further expand this by emphasizing technical principles like fairness, interpretability, and explainability, alongside organizational principles such as transparency, accountability, and benevolence. While these principles offer a comprehensive ethical blueprint, implementing them effectively in real-world systems remains a considerable challenge. To navigate these challenges, we propose a \textit{tiered approach} that establishes progressively stricter \textit{levels} of compliance with ethical principles. Each level represents a set of actionable system properties derived from ethical principles, ensuring they are translated into concrete requirements. Systems must satisfy the foundational properties of lower levels before progressing to higher ones. This structure enables a practical approach to balancing ethical integrity with technological feasibility while accommodating gradual improvements in system capabilities.

\textbf{Level 1} establishes a baseline set of system properties required for safe, minimal-risk usage. Given the inherent risk of generating misleading or factually incorrect content \cite{Jeblick2024, Levine2023.01.30.23285067, birkun_large_2023}, the first Level 1 requirement focuses on the principle of \textit{nonmaleficence}. Additionally, ensuring the confidentiality and security of patient data is critical, as emphasized by Ong et al. \cite{Ong2024}, to uphold patient autonomy and system dependability, both central to Level 1 requirements. These safeguards can be achieved through strict data management protocols and local deployments that minimize data leakage. Robust anonymization techniques \cite{Wiest2024.06.11.24308355} play a key role in securing patient privacy when using data externally.

Systems that comply with this level may be acceptable in lower-risk contexts, where errors can be quickly identified and corrected by qualified professionals. Adhering to Level 1 properties enables the ethical use of language models for a variety of tasks, including most classification and information extraction tasks, as well as many summarization and paraphrasing tasks.

\textbf{Level 2} builds upon Level 1 compliance and introduces additional principles: fairness, interpretability, auditability, and knowledge management. These enhanced requirements enable systems to perform more sensitive tasks, such as decision support, while avoiding direct system-patient interactions. However, modern LLMs often lack interpretability and remain susceptible to bias \cite{xllms, gallegos2024bias, fairnessllms}, a challenge further exacerbated by the absence of well-established, widely applicable metrics to measure these properties. This makes the integration of Level 2 principles an open research area. As a result, the deployment of LLMs in high-stakes applications, such as differential diagnosis, triaging, or modeling patient timelines, should generally involve human-in-the-loop supervision to mitigate risks.

\textbf{Level 3} encompasses the principles of beneficence, justice, explainability, and benevolence, which are critical for systems capable of interacting with patients without supervision from healthcare professionals. A system can achieve Level 3 only after fully satisfying the requirements of Levels 1 and 2. Attaining this tier would enable widespread, ethically responsible deployment across all reviewed medical tasks. However, to the best of our knowledge, no current models or systems meet this standard. Table~\ref{tab:ethical_levels} summarizes the proposed hierarchical levels, outlines scenarios of use for models and systems that comply with the corresponding ethical principles adopted from Solomonides et al. \cite{Solomonides}, and provides representative tasks for each level.

\begin{table*}[h]
\footnotesize
    \begin{adjustwidth}{-2.25in}{0in}

\centering
\caption{Levels of Ethical Compliance and Corresponding Scenarios in System or Model Applications}
\begin{tabular}{p{1.5cm} p{4cm} p{6 cm} p{6cm}}
\toprule
\textbf{Level} & \textbf{Ethical Principles Compliance} & \textbf{Scenarios of Use} & \textbf{Sample Tasks} \\
\midrule
\textbf{Level 1} & Nonmaleficence, Autonomy, Dependability & 
- Non-critical tasks with low risk to patients \newline
- Tasks under supervision of professionals \newline 
- Tasks that exclude direct interaction of patients with a system & 
Phenotyping, Medical Coding, Eponyms Disambiguation, Clinical Acronym Disambiguation, Generating Textual Descriptions from Graph Models, Generating Summaries for Patient-Provider Dialogues, Clinical Note Generation, Medical Report Generation, Querying Data from Electronic Health Records, Extracting Information from Clinical Narrative Reports, Extracting Information from Medical Articles, Summarizing Clinical Study Reports, Summarizing Patient-Provider Dialogues, Anonymization of Medical Documents  \\
\midrule
\textbf{Level 2} & Fairness, Interpretability, Auditability, Knowledge Management, Accountability & 
- Tasks that involve decision-making without direct interaction of patients with a system \newline

& 
Triaging, Differential Diagnosis, Social Media Monitoring, Modeling Patient Timeline   \\
\midrule
\textbf{Level 3} & Beneficience, Justice, Explainability, Benevolence  & - Patient-facing applications &
Medical Chatbots and Health Assistants   \\
\bottomrule
\end{tabular}
\label{tab:ethical_levels}
\end{adjustwidth}
\end{table*}

From an ethical standpoint, current language model systems are far from fully ready for deployment in high-stakes clinical settings. While many systems can meet the baseline requirements of Level 1 for low-risk tasks under professional supervision, substantial challenges remain in achieving compliance with higher levels. The lack of widely adopted interpretability, fairness, and accountability metrics hampers progress toward Level 2. To our best knowledge, no existing systems currently meet the stringent requirements of Level 3, which are essential for patient-facing applications.

\color{black}

\subsection{Datasets}
With new applications of textual AI emerging in areas like medication plan generation, triaging, extracting structured data from medical records, and providing medical consultations, the development of novel, open, and de-identified datasets becomes increasingly necessary. Many existing datasets were created before the advent of LLMs, which may inflate study results and lead to an overestimation of the current models' efficacy. Moreover, access to many existing datasets requires special approvals, which hinders widespread research in this area. Future efforts should focus on creating and utilizing open datasets specifically designed to evaluate LLMs in the medical domain to more accurately reflect their true capabilities.

\subsection{Human-AI Collaboration}

\color{black}
\color{black}
Further research is required to enhance our understanding and optimization of human-AI collaboration in healthcare. This includes exploring how medical professionals can best interact with and leverage AI tools for improved decision-making and patient care, as well as reducing routine work to help prevent burnout. An example of this could be the further exploration of AI-in-the-loop agents, similar to those described in \cite{sharma2022humanai}. 
\color{black}

\subsection{Necessity for Empirical Studies}
Empirical research on real-world use-cases of AI in healthcare is essential. Theoretical studies have broadened our understanding, but practical challenges in real healthcare environments, such as hospitals and clinics, are less understood. Research should focus on how AI applications integrate with healthcare systems, their impact on workflows and healthcare professionals, and the long-term effects on patient outcomes, staff efficiency, and costs. Additionally, addressing AI implementation challenges, including data privacy, ethical concerns, and the need for ongoing system training and updates, is vital. This will guide best practices for AI integration, reduce risks, and ensure these technologies effectively enhance patient care and healthcare delivery.

\section{Conclusion}

\color{black}
This study provides an in-depth examination of recent advancements in language models within the medical domain, with a particular emphasis on clinical applications and locally deployable solutions. It traces the development of language models, exploring both general-purpose and domain-specific architectures, and evaluates their role in medical contexts. The study highlights key tasks performed by these models, such as text generation, token classification, and question answering, demonstrating their practical utility through real-world healthcare scenarios.

Recent advancements in the field have introduced more comprehensive approaches, particularly through multimodal models that seamlessly integrate visual and textual data. These innovations enable holistic AI solutions and are bolstered by techniques such as parameter-efficient fine-tuning and flash attention, which significantly reduce computational requirements. The rise of generative LLMs with in-context learning capabilities marks a pivotal evolution, unlocking new possibilities in specialized medical domains like radiology report generation and medical chatbots - tasks that were considered unattainable just a decade ago.

However, deploying language models in healthcare presents several challenges. The first challenge lies in evaluating generative models, especially for complex tasks like medical conversations or summarizations, where task-specific metrics are insufficient, and extrinsic evaluations require intricate frameworks or human alignment. Additionally, critical domain-specific requirements, such as privacy, up-to-dateness, and empathy, are difficult to quantify or standardize in healthcare applications. Lastly, the lack of automated frameworks for clinical impact validation complicates the evaluation process further, hindering the ability to assess real-world outcomes, such as improved patient care or administrative efficiency. 

The second challenge revolves around the ethical deployment of language models in clinical settings. While many systems meet baseline compliance for low-risk tasks under professional supervision, advancing to higher levels of ethical compliance remains demanding. Requirements such as fairness and interpretability are hindered by the absence of widely adopted metrics and the persistent biases in modern LLMs. Fully satisfying the comprehensive set of ethical standards essential for patient-facing applications is particularly daunting, and, to the best of our knowledge, no current systems meet these stringent requirements.

Mitigating these challenges requires both technological advancements and focused research. Data privacy concerns can be addressed through locally deployed models or robust anonymization techniques, while the up-to-dateness requirement can be managed using retrieval-augmented generation techniques for generative models or continuous adaptation for encoder-based models. These advancements make a wide range of tasks, those that exclude direct patient interaction and unsupervised decision-making, technically viable for broader adoption within medical organizations. However, even these tasks may require additional regulatory approvals and evaluation in real-world clinical settings before widespread implementation \cite{SBLENDORIO2024105501}. 

For tasks involving unsupervised decision-making, addressing privacy and model up-to-dateness alone may not suffice. This is especially true for generative models producing unstructured text, as there is a lack of widely adopted automated metrics for assessing both ethical compliance and performance. A practical approach for handling such tasks is to reformulate them into well-defined tasks, such as classification or information extraction, along with incorporating the latest developments in reasoning and interpretability \cite{jie2024interpretable,liu2024two, ong2025explainable}. In these contexts, established task-specific metrics are sufficient to evaluate performance, and mathematical formulations for assessing ethical components like fairness \cite{fairness} are readily available. If the approach with reformulating the tasks is not feasible, research exploring the integration of  LLMs with ontologies, graph attention networks, and other more deterministic and interpretable models represents another promising direction in moving the adoption of models feasible for unsupervised decision-making. 

Finally, we find no existing solutions that are ethically and technically prepared for tasks involving direct interaction with patients beyond purely administrative functions. 

In general, the specific demands of the medical field, where precision is paramount due to the high cost of errors and ethical compliance must address a wide range of principles such as interpretability, fairness, and accountability, result in an unavoidable \textit{temporal gap} between technological advancements and their adoption. To address this gap, future research, besides purely technological advancements, should prioritize empirical studies in real-world settings to explore how AI can be seamlessly integrated into healthcare workflows without increasing the burden on healthcare professionals. Additionally, it should assess the impact of AI on patient care and its long-term implications for outcomes and costs. We also advocate for the development of \textbf{Medical Model Cards}, inspired by the generic Model Cards proposed by Mitchell et al. \cite{mitchell2019model}. These Medical Model Cards should accompany all models intended for use in clinical settings, providing detailed information about compliance with ethical principles, validated through appropriate metrics or benchmarks. They should also include the intended tasks and corresponding performance benchmarks. This framework will facilitate quicker and more informed model selection for adoption in the healthcare domain.

\color{black}

\nolinenumbers

\bibliography{bibliography}

\end{document}